\pdfoutput=1

\documentclass[11pt]{article}

\usepackage[final]{coling}

\usepackage{times}
\usepackage{latexsym}

\usepackage[T1]{fontenc}

\usepackage[utf8]{inputenc}

\usepackage{microtype}

\usepackage{inconsolata}

\usepackage{graphicx}
\usepackage{enumitem} 
\usepackage{amsmath} 
\usepackage{booktabs}
\usepackage{hyperref}
\usepackage{multirow}
\usepackage{tabularx}

\title{EasyJudge: an Easy-to-use Tool for Comprehensive Response Evaluation of LLMs}

\author{
  \textbf{Yijie Li\textsuperscript{1,2}},
  \textbf{Yuan Sun\textsuperscript{1,2,*}}\\
  \textsuperscript{1}Minzu University of China, Beijing, China\\
  \textsuperscript{2}National Language Resource Monitoring \& Research Center Minority Languages Branch\\
  Emails: \texttt{yijie\_li\_cn@163.com}, \texttt{tracy.yuan.sun@gmail.com}\\
  \small{\textbf{* Corresponding author:} Yuan Sun}
}

\begin{document}
\maketitle
\begin{abstract}
Recently, there has been a growing trend of employing large language models (LLMs) to judge the quality of other LLMs. Many studies have adopted closed-source models, mainly using GPT-4 as the evaluator. However, due to the closed-source nature of the GPT-4 model, employing it as an evaluator has resulted in issues including transparency, controllability, and cost-effectiveness. Some researchers have turned to using fine-tuned open-source LLMs as evaluators. However, existing open-source evaluation LLMs generally lack a user-friendly visualization tool, and they have not been optimized for accelerated model inference, which causes inconvenience for researchers with limited resources and those working across different fields. This paper presents EasyJudge, a model developed to evaluate significant language model responses. It is lightweight, precise, efficient, and user-friendly, featuring an intuitive visualization interface for ease of deployment and use. EasyJudge uses detailed datasets and refined prompts for model optimization, achieving strong consistency with human and proprietary model evaluations. The model optimized with quantitative methods enables EasyJudge to run efficiently on consumer-grade GPUs or even CPUs. We also provide detailed analysis and case studies to further reveal the potential of our method. \footnote{Code is open at \href{https://github.com/4real3000/EasyJudge}{https://github.com/4real3000/EasyJudge}. Video demonstrations at \href{https://youtu.be/3NcSWPf9rzM}{https://youtu.be/3NcSWPf9rzM}.}

.

\end{abstract}

\section{Introduction}

The evaluation of response quality from large language models (LLMs) has been a central concern within the research community\citep{liang2022holistic,chang2024survey}. As the instruction-following capabilities of LLMs continue to evolve, a more comprehensive and precise evaluation of their responses becomes particularly crucial\citep{qin2023chatgpt}. Traditional evaluation metrics such as BLEU\citep{papineni2002bleu}, ROUGE\citep{lin2004rouge}, BERTScore\citep{zhang2019bertscore}, BARTScore\citep{yuan2021bartscore}, and GPTScore\citep{fu2023gptscore} primarily offer shallow semantic analysis and assessment for basic natural language processing tasks. Due to their limited scope and poor interpretability, traditional metrics are ill-suited for the demands of large language models, especially as tasks evolve to better align with human needs. 

Some studies have proposed the concept of LLM-as-a-Judge\citep{li2023alpacaeval,zheng2023judging}, which leverages proprietary LLMs, particularly GPT-4\citep{achiam2023gpt}, to evaluate the responses of other LLMs. By defining evaluation schemes within prompts, LLMs can utilize their instruction-following capabilities to provide reliable assessments, achieving high consistency with human evaluators. However, relying on external APIs for evaluation raises potential privacy concerns, and the lack of transparency in API models poses challenges to the reproducibility of the evaluations. Moreover, using APIs can result in significant cost overhead. For instance, evaluating four different LLM variants (ranging from 7B to 65B in size) across 1,000 evaluation instances using GPT-4 could exceed \$2,000. Such costs are often prohibitive for academic institutions or researchers operating under limited budgets\citep{kim2023prometheus}.

A mainstream alternative approach is to train a evaluation model based on open-source LLMs. For example, PandaLM\citep{wang2023pandalm} and JudgeLM\citep{zhu2023judgelm} construct datasets from diverse instruction sets and annotations from GPT-series models, fine-tuning open-source models like LLaMA\citep{touvron2023llama} to serve as scalable evaluation models. Auto-J\citep{li2023generative} and Prometheus\citep{kim2023prometheus} explore the refinement of model evaluation metrics, aiming to build fine-grained evaluation models.

However, current LLM-as-Judge research typically provides only a fine-tuned LLM, lacking an user-friendly visualization interface tailored for LLM evaluation. This poses challenges for users who seek a one-stop, simple, and efficient solution to evaluate responses generated by some models. 




To advance large model evaluation in routine research, this study aims to develop an LLM-as-Judge evaluation model and platform named EasyJudge. EasyJudge can evaluate model responses using POINTWISE (direct scoring) and PAIRWISE (pairwise ranking). It is fine-tuned on a dataset of real-world LLM instruction responses, carefully categorized into 50 scenario types, with responses from over ten open-source LLMs in the training data. EasyJudge provides two methods for evaluating model responses: POINTWISE (direct scoring) and PAIRWISE (pairwise ranking). The model is fine-tuned on a carefully curated dataset of real-world LLM instruction responses, organized into 50 distinct scenario categories, and includes response data from over ten open-source LLMs.

Additionally, we defined 8-10 specific evaluation criteria for each of the 50 scenario categories, resulting in 139 evaluation criteria related to LLM responses. In this work, these multi-scenario, multi-criteria instruction datasets were used to fine-tune the LLaMA-3-8b model. The fine-tuned model can provide precise and multidimensional evaluations of LLMs’ rather than generalized assessments. Additionally, this work employs techniques such as quantization and mixed precision to reduce memory usage and resource overhead during runtime, thereby achieving faster inference speeds. Finally, the model has been encapsulated to provide users with a simplified, user-friendly interface that is clear and intuitive to operate.

The specific features of EasyJudge are as follows:  

\begin{enumerate}
    \item[(1)] Lightweight usage model. EasyJudge is built to minimize dependency requirements, offering a simple installation process and precise documentation. Users can initiate the evaluation interface with only a few basic commands.
    \item[(2)] Comprehensive evaluation tool. EasyJudge offers a highly customizable interface, allowing users to select evaluation scenarios and flexibly combine evaluation criteria based on their needs. The visualization interface has been carefully designed to provide users with an intuitive view of various aspects of the evaluation results.
    \item[(3)] Efficient inference engine. EasyJudge employs model quantization, memory management optimization, and hardware acceleration support to enable efficient inference. As a result, EasyJudge can run seamlessly on consumer-grade GPUs and even CPUs.
\end{enumerate}

\begin{figure*}[htbp]
    \centering
    \includegraphics[width=\textwidth]{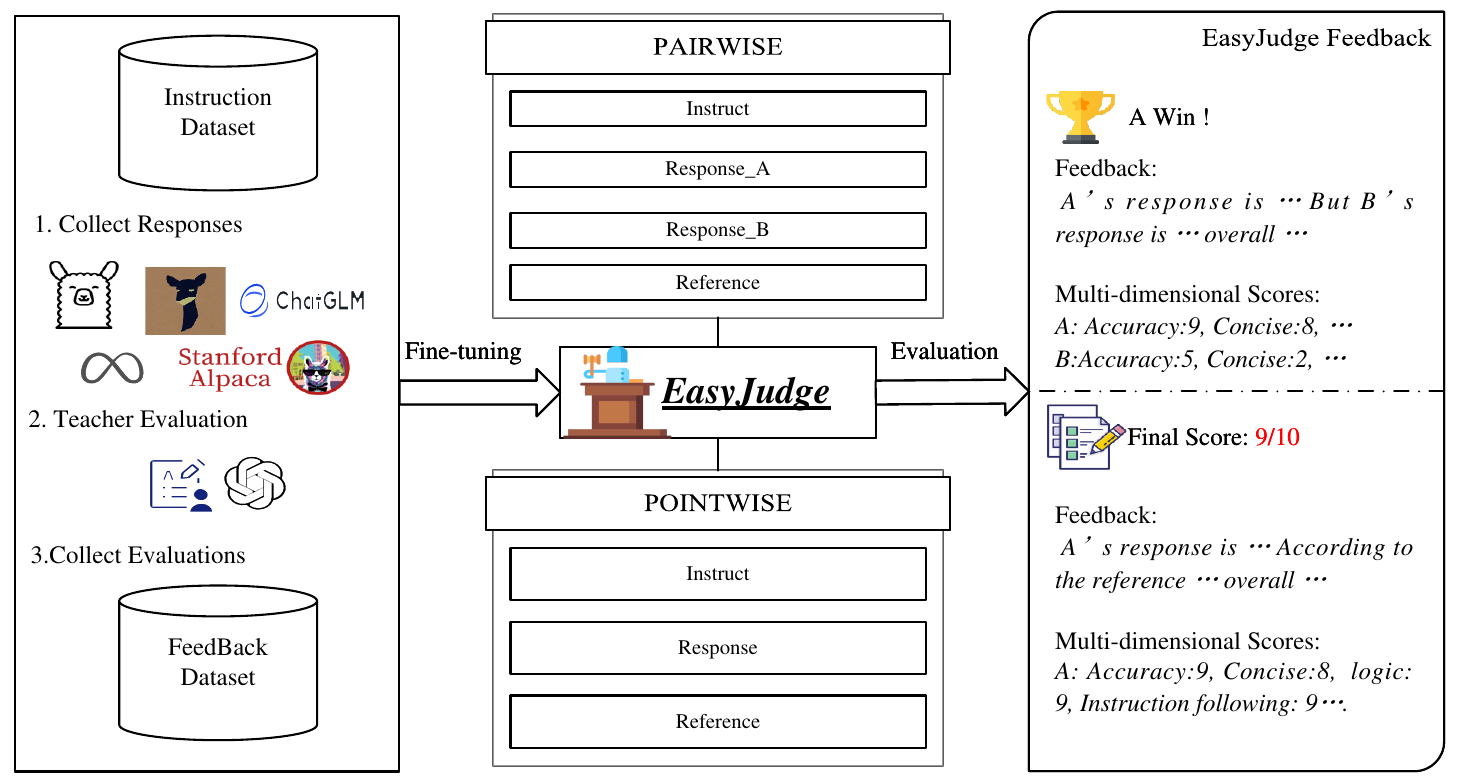}
    \caption{Overview of the EasyJudge method.}
    \label{fig:overview}
\end{figure*}

\section{Related Work}
\subsection{Evaluation Based on Reference Texts}
Traditional model-free scoring methods like BLEU\citep{papineni2002bleu} and ROUGE\citep{lin2004rouge} were widely used but have limitations in evaluation reliability. Recent model-based methods, such as BERTScore\citep{zhang2019bertscore}, BLEURT\citep{sellam2020bleurt}, and BARTScore\citep{yuan2021bartscore}, improve evaluation by capturing semantic-level information. EasyJudge visually compares responses using metrics like ROUGE, BLEU, and BERTScore, offering users a more comprehensive and intuitive evaluation of models.

\subsection{LLM-Based Text Evaluation}
Recent research has shifted towards using large language models as evaluators, employing GPT-4 or fine-tuned Judge LLMs to assess the text quality generated by other models. Recent studies have shown that ChatGPT can outperform crowdsourced workers in text annotation tasks\citep{gilardi2023chatgpt,chiang2023can}.Using closed-source models like GPT-4 for evaluation poses challenges, including high costs, privacy risks, and limited control. Fine-tuned open-source Judge LLMs, such as PandaLM\citep{wang2023pandalm}, AUTO-J\citep{li2023generative}, PROMETHEUS\citep{kim2023prometheus}, JudgeLM\citep{zhu2023judgelm}, and Eval-Instruct\citep{wu2024instructeval}, have been developed to overcome this. These models offer cost-effective, reliable evaluation solutions, addressing issues like data leakage, evaluation bias and adapting to diverse tasks. They collectively advance LLM evaluation by integrating subjective criteria, enhancing multimodal and dialogue tasks, and providing alternatives to closed-source models.

However, current LLM-as-Judge research typically only provides fine-tuned LLMs, lacking an intuitive and user-friendly visualization interface specifically optimized for LLM evaluation. This presents challenges for users who seek a simple and efficient one-stop solution for evaluating individual responses or entire texts. Additionally, users are unable to intuitively access evaluation results from these models. A comparison between EasyJudge and these evaluation models is provided in \hyperref[tab:comparison]{Table \ref*{tab:comparison}}.

\begin{table*}[h]
\centering
\resizebox{\textwidth}{!}{
\begin{tabular}{lcccccc}
\toprule
Name        & Foundation     & Evaluation scheme     & Web GUI   & Result visualization & Inference acceleration \\
\midrule
PandaLM\citep{wang2023pandalm}     & LLaMA          & Pairwise              & Yes       & No                   & No                     \\
JudgeLM\citep{zhu2023judgelm}     & Vicuna         & Pairwise              & Yes       & No                   & No                     \\
Auto-J\citep{li2023generative}      & LLaMA2-chat    & Pairwise/Pointwise    & No        & No                   & No                     \\
Prometheus\citep{kim2023prometheus}  & LLaMA2-chat    & Pointwise             & No        & No                   & No                     \\
EasyJudge(ours)   & LLaMA3-instruct& Pairwise/Pointwise    & \textbf{Yes}       & \textbf{Yes}                  & \textbf{Yes}                    \\
\bottomrule
\end{tabular}
}
\caption{Comparison of Response Evaluation Methods Based on LLMs.}
\label{tab:comparison} 
\end{table*}

\section{System Overview}

This section provides a detailed overview of the EasyJudge system. As shown in \hyperref[fig:overview]{Figure 1}, EasyJudge consists of three key components:


\subsection{Data Processing Module}
We collect real-world interaction data between humans and Large Language Models to create the initial Instruction Dataset. A classifier is then trained to categorize this instruction data. Additionally, GPT-4 is employed to expand the instructions through prompts. Multiple open-source large models are then used to generate responses to the instruction data. Finally, GPT-4 is invoked with carefully designed prompts that include detailed evaluation criteria to produce evaluation results. The data is then integrated for use in the subsequent model fine-tuning.

\subsection{Evaluation Model Training}
This is the core of EasyJudge, where the LLaMA-3-8b base model is fine-tuned using the multi-scenario, multi-criteria instruction data obtained from the data processing phase. The result is the POINTWISE model for direct evaluation and the PAIRWISE model for pairwise comparison evaluation. Next, model merging techniques are applied to integrate the performance of both models. Finally, model quantization techniques are used to optimize the model. 

\subsection{User-Friendly Interface}
The evaluation process of model responses is designed to be transparent, offering users an intuitive interface with several key features. These include selecting models, adjusting model parameters, configuring evaluation scenarios, and customizing evaluation criteria. Additionally, the interface provides a clear visualization of the evaluation results. For example, it displays the outcomes of pairwise comparisons and direct scoring in a straightforward manner, offers detailed feedback, and presents multi-dimensional score references to help users better understand the evaluation process.

\section{Implementation Details}

To better understand the evaluation process within EasyJudge, this section will explain the implementation of three key issues.

\subsection{Data Processing}
\subsubsection{Definition of Evaluation Scenarios and Criteria}
To ensure a more accurate and context-relevant evaluation of LLM responses, EasyJudge, based on prior research\citep{kim2023prometheus,zhu2023judgelm,li2023generative}, categorizes evaluation scenarios into 50 distinct types, which are further summarized into nine broader categories: text generation and writing, information extraction and analysis, mathematics and logical reasoning, code tasks, QA, reasoning and judgment, role-playing and conversation, basic NLP tasks, and a default type. It is intuitive to understand that the evaluation criteria for responses in different scenarios, such as code generation and writing a project proposal, should vary significantly. Therefore, EasyJudge customizes evaluation criteria for each of the 50 distinct scenarios. Each category includes 8-10 evaluation criteria, totalling 134 unique criteria divided into four main categories: Basic, Style, Content, and Format.

\subsubsection{Dataset Construction}
High-quality datasets are crucial for effectively fine-tuning large language models to serve as evaluation judges. However, existing datasets and prior research often lack sufficient diversity and detailed evaluation criteria. To address these issues, EasyJudge introduces a new dataset that includes various seed tasks across different evaluation scenarios, comprehensive answers from multiple open-source LLMs, scoring results from a teacher LLM across various criteria dimensions, and detailed reasoning behind each evaluation.

EasyJudge extracts 15k seed tasks from a large-\_qa, flan, truthful\_qa, and ultrachat. A classification model is then used to categorize the instructions based on the scenario definitions described in section 4.1.1. For scenarios with limited instructions, GPT-4 is employed to supplement them using the self-instruct method. The prompt template used for this process is provided in \hyperref[fig:The prompt for invoking GPT-4 extended instructions.]{Figure 3}. To enhance the diversity of the dataset, we aggregate responses from multiple open-source LLMs, including but not limited to LLaMA, Alpaca, and Vicuna. Next, we combine the LLM-generated responses with reference answers to create an answer set. For PAIRWISE tasks, two responses from different open-source models are randomly selected from the answer set for the same instruction. An advanced teacher model, GPT-4, is then used to assign detailed scores and provide thorough reasoning for the comparison. For POINTWISE tasks, a response from an open-source model is randomly selected from the answer set for a given instruction. The advanced teacher model, GPT-4, then assigns detailed scores and provides comprehensive reasoning for the evaluation. To ensure robust and comprehensive judgments, we utilized detailed prompt templates, the specifics of which are provided in \hyperref[fig:The prompt used for PAIRWISE instruction evaluation.]{Figure 4} and \hyperref[fig:The prompt used for POINTWISE instruction evaluation.]{Figure 5}. The prompt contains critical inputs such as the scenario, evaluation criteria, instruction, and response to be evaluated, along with the evaluation requirements and output format. Including these details ensures the model produces clear, comprehensive, and accurate evaluation results.

\subsection{Evaluation Model Fine-Tuning}
The data required to train the EasyJudge model is constructed by integrating the datasets mentioned in section 4.1.2. The training data follows the Alpaca fine-tuning format, which consists of four components: instruction, input, output, and system. The instruction includes the task to be evaluated and its corresponding response, evaluation requirements and the output format; the input is left empty by default, the output contains the scores and reasoning provided by the teacher model GPT-4, and the system includes the scenario and evaluation criteria. The data templates are in \hyperref[fig:ALPACA fine-tuning data template.]{Figure 6}.

To reduce positional bias in PAIRWISE comparisons, EasyJudge applies a simple data augmentation technique. According to the judgelm, For each pairwise training sample, the order of the two responses in the input is randomly swapped. Additionally, to enhance the model's ability to handle unknown responses, EasyJudge randomly drops the reference for each data point\citep{zhu2023judgelm}.

EasyJudge adopts the LLaMA-3-8b model as its base LLM and utilizes the LLaMA-Factory framework for model fine-tuning. Training parameter details can be found in \hyperref[tab:training_params]{Table \ref*{tab:training_params}}. The PAIRWISE evaluation model is fine-tuned using 5k data points, while the POINTWISE evaluation model uses 10k data points.

Moreover, EasyJudge employs the DARE weight merging strategy to integrate models trained under different evaluation modes while applying INT8 quantization to significantly reduce model size and inference time, enhancing deployment efficiency and applicability without compromising evaluation performance.

To demonstrate the superior performance of the EasyJudge model in evaluation tasks, this paper presents the model's test results on the PandaLM-test and Prometheus-test-ood datasets. Specifically, the PandaLM-test dataset is used to assess the performance of the pairwise model, while the Prometheus-test-ood dataset is employed to evaluate the pointwise model. The results are shown in \hyperref[tab:results]{Table \ref*{tab:results}}. Further test results will be discussed in future work.

\begin{table}[h]
\centering
\resizebox{\columnwidth}{!}{%
\begin{tabular}{lccccc}
\toprule
\multirow{2}{*}{Model}   & \multicolumn{2}{c}{PandaLM-test} & \multicolumn{2}{c}{Prometheus-test-ood} \\
                         & Accuracy      & F1              & Pearson       & Spearman \\
\midrule
JudgeLM-7B               & 70.97         & 67.59           & 0.610         & 0.690 \\
PandLM-7B                & 67.57         & 57.49           & 0.386         & 0.383 \\
Auto-J-13B               & 71.47         & 61.01           & 0.591         & 0.580 \\
EasyJudge-8B             & \textbf{71.83}         & \textbf{68.36}           & \textbf{0.679}         & \textbf{0.701} \\
\bottomrule
\end{tabular}
}
\caption{Results of evaluators on pairwise selection and pointwise grading.}
\label{tab:results}
\end{table}

\subsection{User-Friendly Interface Development}
To make the evaluation process of large language models more intuitive and user-friendly, we developed a Streamlit-based interface. Using the Streamlit framework, we created a transparent and responsive interface. This interface not only supports data upload and parameter adjustments but also allows users to select evaluation scenarios dynamically through radio buttons. The evaluation results are presented in a rich format, including text and graphical representations, ensuring that users can easily interpret the meaning of the model's output. This design significantly reduces the operational complexity of EasyJudge and enhances user experience, enabling even non-expert users to perform advanced model evaluations efficiently. This intuitive interface and the model's efficient computation capabilities can meet diverse evaluation needs, particularly in resource-constrained environments.

\section{Demonstration Scenarios}

\begin{figure*}[htbp]
    \centering
    \includegraphics[width=\textwidth]{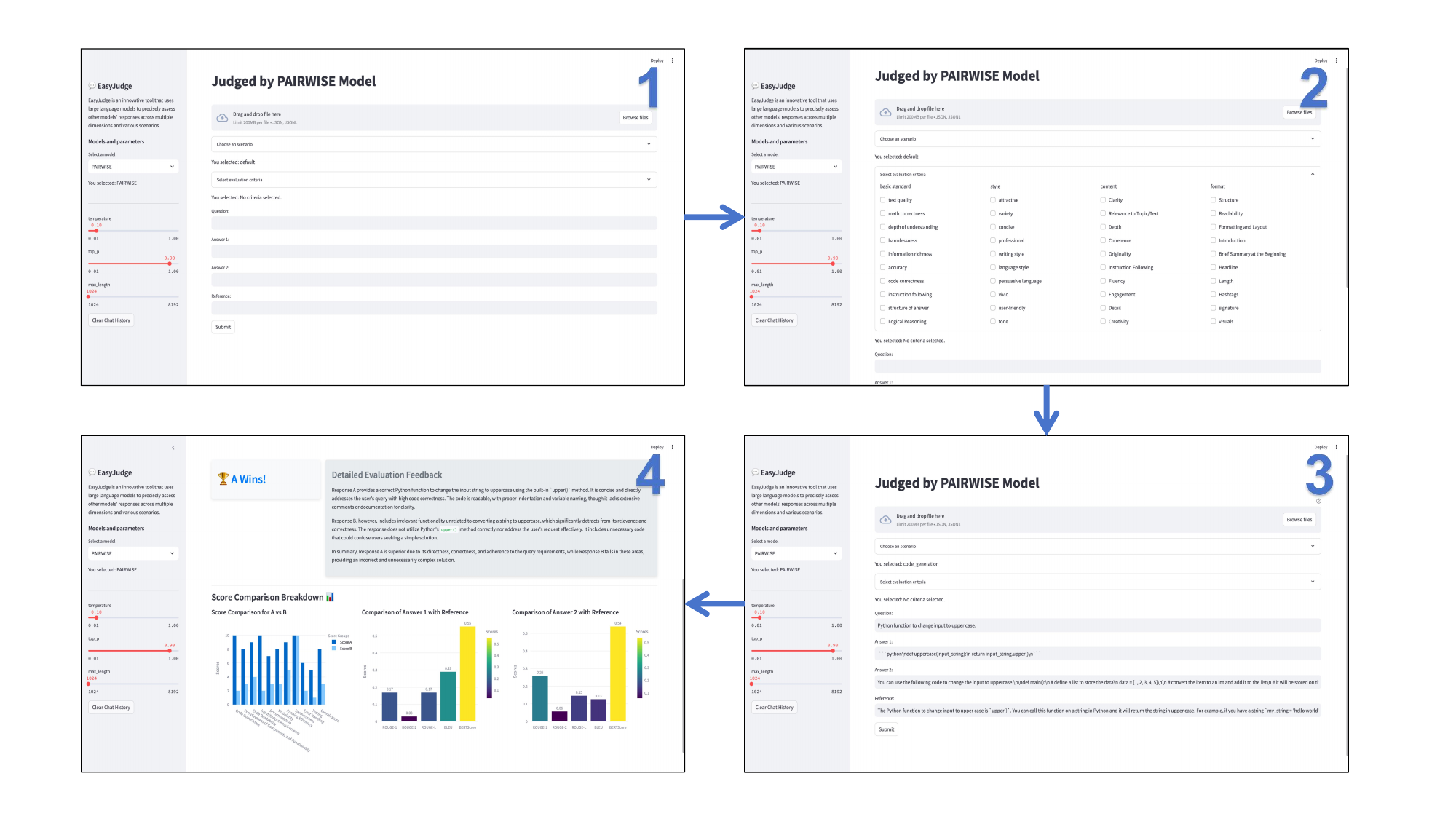}
    \caption{A Screenshot of EasyJudge with an example evaluation task of PAIRWISE.}
    \label{fig:screenshot}
\end{figure*}
\subsection{Diversity Classification}
\hyperref[fig:screenshot]{Figure 2} provides a screenshot of the EasyJudge user interface, through which users can perform large model response evaluations by following these steps:

\textbf{Step 1 (Task Configuration)}: As shown in \hyperref[fig:screenshot]{Figure 2-1}, the configuration interface guides users through the initial setup. Users begin by selecting the evaluation task type, which includes single response direct scoring (POINTWISE) and pairwise response ranking (PAIRWISE). The system automatically selects the appropriate prompt template based on the chosen scoring strategy. After selecting the task, another essential configuration allows users to adjust the EasyJudge model parameters, such as temperature, Top-p, and max\_length, according to their specific needs to achieve optimal evaluation results.

\textbf{Step 2 (Scenario and Criteria Configuration)}: Next, users can select specific task scenarios and criteria tailored to their evaluation data, a crucial advantage of EasyJudge's highly customizable system. The evaluation criteria configuration is shown in \hyperref[fig:screenshot]{Figure 2-2} (scenario configuration can be found in \hyperref[fig:user_scenario]{Figure 7}). Once users select a task scenario, EasyJudge conducts a multi-dimensional evaluation based on the specific criteria. Alternatively, users can opt not to select a scenario where EasyJudge will evaluate the model response using the default scenario. The default scenario encompasses ten standard evaluation criteria suitable for most task evaluations. For more customized evaluation results, users can manually select the criteria. EasyJudge currently offers 40 evaluation criteria across four main categories for tailored evaluation. If custom criteria are selected, the system will automatically bypass scenario selection.

\textbf{Step 3 (Data Upload)}: As shown in \hyperref[fig:screenshot]{Figure 2-3}, EasyJudge provides two methods for data upload. Suppose a user is evaluating a single data instance. In that case, they can sequentially copy the instruction, Model 1's response, Model 2's response, and the reference answer into the corresponding input fields, then click the submit button to initiate the evaluation. To evaluate multiple data instances, users must upload a JSON/JSONL file containing the evaluation data in Alpaca format. After uploading the file, users can click the submit button to start the evaluation process. The interface for single data evaluation in the POINTWISE mode is shown in \hyperref[fig:user_POINTWISE_1]{Figure 8}.

\textbf{Step 4 (Results Display)}: In this step, EasyJudge presents the final evaluation results, providing users with clear evaluation information that helps them intuitively understand the quality of the responses. Taking PAIRWISE evaluation as an example (Details on POINTWISE evaluation can be found in Appendix \ref{sec:appendix-c}.), as shown in \hyperref[fig:screenshot]{Figure 2-4}, the top of the page displays the final evaluation results. At the same time, the middle section presents the Detailed Evaluation Feedback from the EasyJudge model. At the bottom, three charts (labeled as Figures A, b, and c) are provided: Figure A shows the scores of Response A and Response B across different evaluation criteria dimensions and compares the two responses in each dimension. Figure b compares Response A with the reference text, displaying evaluation results based on traditional metrics, including ROUGE, BLEU, BERTScore, BLEURT, and BARTScore. Figure c compares Response B with the reference answer evaluated using traditional metrics. Finally, if users choose to upload a JSON/JSONL file, then they can finally download a JSON file containing the evaluation result data by clicking the "Download" button.

\section{Conclusion}
This paper introduces EasyJudge, an innovative tool for evaluating LLMs with advanced models, offering a customizable interface and precise, multi-dimensional assessments. It enhances efficiency through model quantization, enabling use on consumer-grade GPUs and CPUs. Future research plans include integrating new technologies to extend EasyJudge's capabilities to evaluate multimodal models, Retrieval-Augmented Generation (RAG), and intelligent agents, contributing to advancing Artificial General Intelligence (AGI) while improving evaluation accuracy and practicality across diverse scenarios.

\section*{Acknowledgments}
This work is supported by the National Nature Science Foundation (61972436), the National Social Science  Foundation (22\&ZD035), and
the Minzu University of China Foundation (GRSCP202316, 2023QNYL22, 2024GJYY43).

\newpage
\bibliography{custom}

\begin{thebibliography}{20}
\providecommand{\natexlab}[1]{#1}

\bibitem[{Achiam et~al.(2023)Achiam, Adler, Agarwal, Ahmad, Akkaya, Aleman, Almeida, Altenschmidt, Altman, Anadkat et~al.}]{achiam2023gpt}
Josh Achiam, Steven Adler, Sandhini Agarwal, Lama Ahmad, Ilge Akkaya, Florencia~Leoni Aleman, Diogo Almeida, Janko Altenschmidt, Sam Altman, Shyamal Anadkat, et~al. 2023.
\newblock Gpt-4 technical report.
\newblock \emph{arXiv preprint arXiv:2303.08774}.

\bibitem[{Chang et~al.(2024)Chang, Wang, Wang, Wu, Yang, Zhu, Chen, Yi, Wang, Wang et~al.}]{chang2024survey}
Yupeng Chang, Xu~Wang, Jindong Wang, Yuan Wu, Linyi Yang, Kaijie Zhu, Hao Chen, Xiaoyuan Yi, Cunxiang Wang, Yidong Wang, et~al. 2024.
\newblock A survey on evaluation of large language models.
\newblock \emph{ACM Transactions on Intelligent Systems and Technology}, 15(3):1--45.

\bibitem[{Chiang and Lee(2023)}]{chiang2023can}
Cheng-Han Chiang and Hung-yi Lee. 2023.
\newblock Can large language models be an alternative to human evaluations?
\newblock \emph{arXiv preprint arXiv:2305.01937}.

\bibitem[{Fu et~al.(2023)Fu, Ng, Jiang, and Liu}]{fu2023gptscore}
Jinlan Fu, See-Kiong Ng, Zhengbao Jiang, and Pengfei Liu. 2023.
\newblock Gptscore: Evaluate as you desire.
\newblock \emph{arXiv preprint arXiv:2302.04166}.

\bibitem[{Gilardi et~al.(2023)Gilardi, Alizadeh, and Kubli}]{gilardi2023chatgpt}
Fabrizio Gilardi, Meysam Alizadeh, and Ma{\"e}l Kubli. 2023.
\newblock Chatgpt outperforms crowd workers for text-annotation tasks.
\newblock \emph{Proceedings of the National Academy of Sciences}, 120(30):e2305016120.

\bibitem[{Kim et~al.(2023)Kim, Shin, Cho, Jang, Longpre, Lee, Yun, Shin, Kim, Thorne et~al.}]{kim2023prometheus}
Seungone Kim, Jamin Shin, Yejin Cho, Joel Jang, Shayne Longpre, Hwaran Lee, Sangdoo Yun, Seongjin Shin, Sungdong Kim, James Thorne, et~al. 2023.
\newblock Prometheus: Inducing fine-grained evaluation capability in language models.
\newblock In \emph{The Twelfth International Conference on Learning Representations}.

\bibitem[{Li et~al.(2023{\natexlab{a}})Li, Sun, Yuan, Fan, Zhao, and Liu}]{li2023generative}
Junlong Li, Shichao Sun, Weizhe Yuan, Run-Ze Fan, Hai Zhao, and Pengfei Liu. 2023{\natexlab{a}}.
\newblock Generative judge for evaluating alignment.
\newblock \emph{arXiv preprint arXiv:2310.05470}.

\bibitem[{Li et~al.(2023{\natexlab{b}})Li, Zhang, Dubois, Taori, Gulrajani, Guestrin, Liang, and Hashimoto}]{li2023alpacaeval}
Xuechen Li, Tianyi Zhang, Yann Dubois, Rohan Taori, Ishaan Gulrajani, Carlos Guestrin, Percy Liang, and Tatsunori~B Hashimoto. 2023{\natexlab{b}}.
\newblock Alpacaeval: An automatic evaluator of instruction-following models.

\bibitem[{Liang et~al.(2022)Liang, Bommasani, Lee, Tsipras, Soylu, Yasunaga, Zhang, Narayanan, Wu, Kumar et~al.}]{liang2022holistic}
Percy Liang, Rishi Bommasani, Tony Lee, Dimitris Tsipras, Dilara Soylu, Michihiro Yasunaga, Yian Zhang, Deepak Narayanan, Yuhuai Wu, Ananya Kumar, et~al. 2022.
\newblock Holistic evaluation of language models.
\newblock \emph{arXiv preprint arXiv:2211.09110}.

\bibitem[{Lin(2004)}]{lin2004rouge}
Chin-Yew Lin. 2004.
\newblock Rouge: A package for automatic evaluation of summaries.
\newblock In \emph{Text summarization branches out}, pages 74--81.

\bibitem[{Papineni et~al.(2002)Papineni, Roukos, Ward, and Zhu}]{papineni2002bleu}
Kishore Papineni, Salim Roukos, Todd Ward, and Wei-Jing Zhu. 2002.
\newblock Bleu: a method for automatic evaluation of machine translation.
\newblock In \emph{Proceedings of the 40th annual meeting of the Association for Computational Linguistics}, pages 311--318.

\bibitem[{Qin et~al.(2023)Qin, Zhang, Zhang, Chen, Yasunaga, and Yang}]{qin2023chatgpt}
Chengwei Qin, Aston Zhang, Zhuosheng Zhang, Jiaao Chen, Michihiro Yasunaga, and Diyi Yang. 2023.
\newblock Is chatgpt a general-purpose natural language processing task solver?
\newblock \emph{arXiv preprint arXiv:2302.06476}.

\bibitem[{Sellam et~al.(2020)Sellam, Das, and Parikh}]{sellam2020bleurt}
Thibault Sellam, Dipanjan Das, and Ankur~P Parikh. 2020.
\newblock Bleurt: Learning robust metrics for text generation.
\newblock \emph{arXiv preprint arXiv:2004.04696}.

\bibitem[{Touvron et~al.(2023)Touvron, Lavril, Izacard, Martinet, Lachaux, Lacroix, Rozi{\`e}re, Goyal, Hambro, Azhar et~al.}]{touvron2023llama}
Hugo Touvron, Thibaut Lavril, Gautier Izacard, Xavier Martinet, Marie-Anne Lachaux, Timoth{\'e}e Lacroix, Baptiste Rozi{\`e}re, Naman Goyal, Eric Hambro, Faisal Azhar, et~al. 2023.
\newblock Llama: Open and efficient foundation language models.
\newblock \emph{arXiv preprint arXiv:2302.13971}.

\bibitem[{Wang et~al.(2023)Wang, Yu, Zeng, Yang, Wang, Chen, Jiang, Xie, Wang, Xie et~al.}]{wang2023pandalm}
Yidong Wang, Zhuohao Yu, Zhengran Zeng, Linyi Yang, Cunxiang Wang, Hao Chen, Chaoya Jiang, Rui Xie, Jindong Wang, Xing Xie, et~al. 2023.
\newblock Pandalm: An automatic evaluation benchmark for llm instruction tuning optimization.
\newblock \emph{arXiv preprint arXiv:2306.05087}.

\bibitem[{Wu et~al.(2024)Wu, Li, Xiao, Liu, and Li}]{wu2024instructeval}
Wenhao Wu, Wei Li, Xinyan Xiao, Jiachen Liu, and Sujian Li. 2024.
\newblock Instructeval: Instruction-tuned text evaluator from human preference.
\newblock In \emph{Findings of the Association for Computational Linguistics ACL 2024}, pages 13462--13474.

\bibitem[{Yuan et~al.(2021)Yuan, Neubig, and Liu}]{yuan2021bartscore}
Weizhe Yuan, Graham Neubig, and Pengfei Liu. 2021.
\newblock Bartscore: Evaluating generated text as text generation.
\newblock \emph{Advances in Neural Information Processing Systems}, 34:27263--27277.

\bibitem[{Zhang et~al.(2019)Zhang, Kishore, Wu, Weinberger, and Artzi}]{zhang2019bertscore}
Tianyi Zhang, Varsha Kishore, Felix Wu, Kilian~Q Weinberger, and Yoav Artzi. 2019.
\newblock Bertscore: Evaluating text generation with bert.
\newblock \emph{arXiv preprint arXiv:1904.09675}.

\bibitem[{Zheng et~al.(2023)Zheng, Chiang, Sheng, Zhuang, Wu, Zhuang, Lin, Li, Li, Xing et~al.}]{zheng2023judging}
Lianmin Zheng, Wei-Lin Chiang, Ying Sheng, Siyuan Zhuang, Zhanghao Wu, Yonghao Zhuang, Zi~Lin, Zhuohan Li, Dacheng Li, Eric Xing, et~al. 2023.
\newblock Judging llm-as-a-judge with mt-bench and chatbot arena.
\newblock \emph{Advances in Neural Information Processing Systems}, 36:46595--46623.

\bibitem[{Zhu et~al.(2023)Zhu, Wang, and Wang}]{zhu2023judgelm}
Lianghui Zhu, Xinggang Wang, and Xinlong Wang. 2023.
\newblock Judgelm: Fine-tuned large language models are scalable judges.
\newblock \emph{arXiv preprint arXiv:2310.17631}.

\end{thebibliography}

\appendix

\section{Prompt Templates}
\begin{figure*}[htbp]
    \centering
    \includegraphics[width=\textwidth]{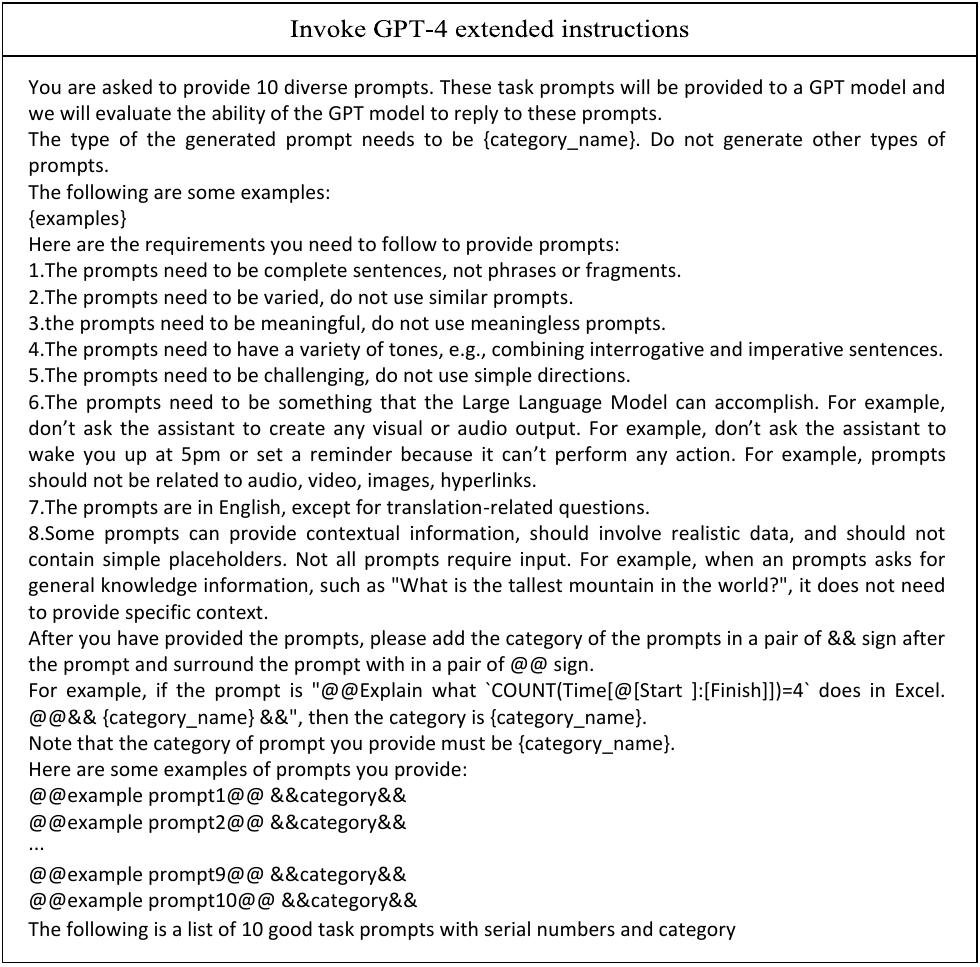}
    \caption{The prompt for invoking GPT-4 extended instructions.}
    \label{fig:The prompt for invoking GPT-4 extended instructions.}
\end{figure*}

\begin{figure*}[htbp]
    \centering
    \includegraphics[width=\textwidth]{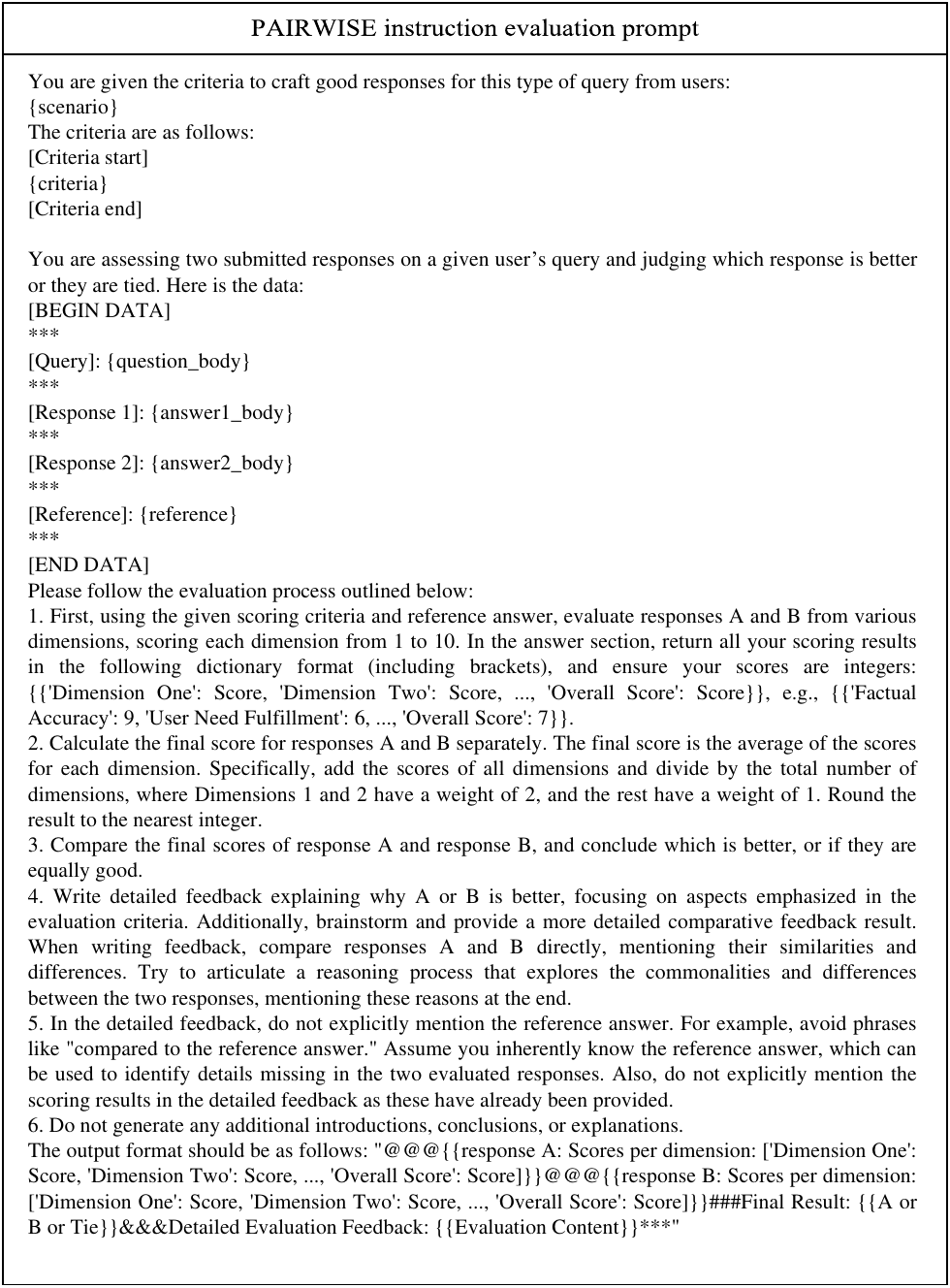}
    \caption{The prompt used for PAIRWISE instruction evaluation.}
    \label{fig:The prompt used for PAIRWISE instruction evaluation.}
\end{figure*}

\begin{figure*}[htbp]
    \centering
    \includegraphics[width=\textwidth]{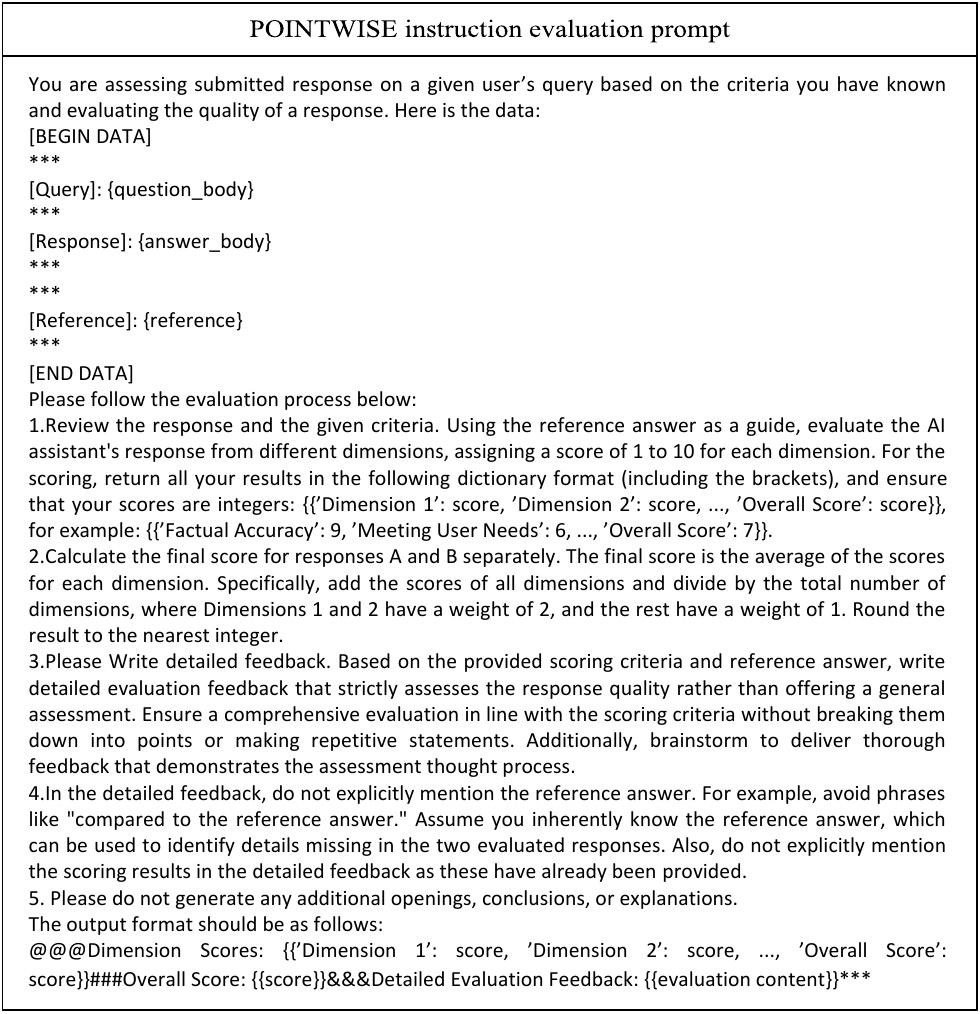}
    \caption{The prompt used for POINTWISE instruction evaluation.}
    \label{fig:The prompt used for POINTWISE instruction evaluation.}
\end{figure*}

\begin{figure*}[htbp]
    \centering
    \includegraphics[width=\textwidth]{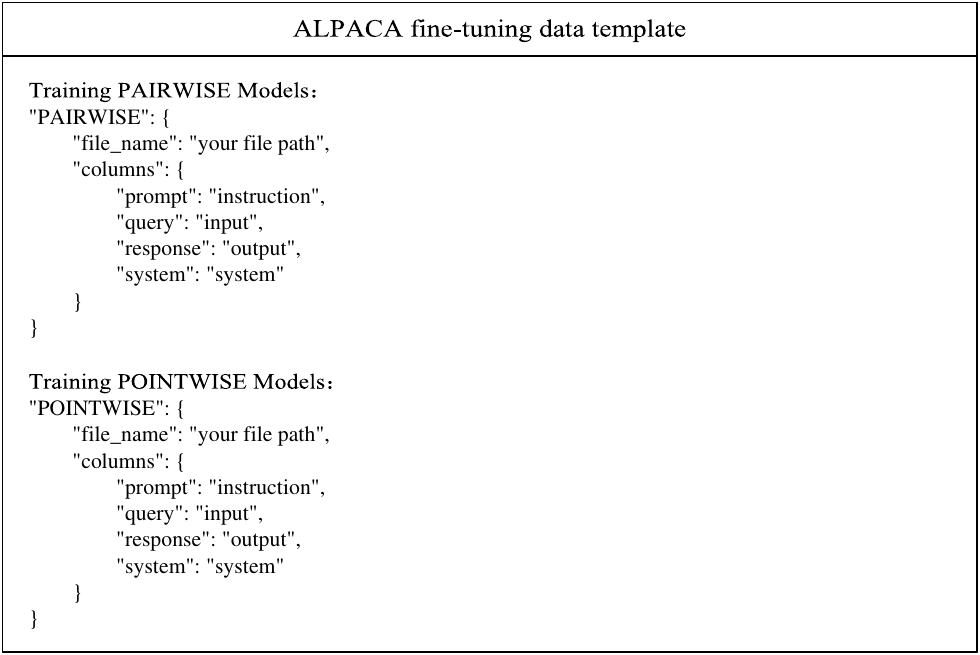}
    \caption{ALPACA fine-tuning data template.}
    \label{fig:ALPACA fine-tuning data template.}
\end{figure*}

This section lists all the prompt templates that EasyJudge used. \hyperref[fig:The prompt for invoking GPT-4 extended instructions.]{Figure 3} shows the prompt for invoking GPT-4 extended instructions. \hyperref[fig:The prompt used for PAIRWISE instruction evaluation.]{Figure 4} shows the prompt used for PAIRWISE instruction evaluation. \hyperref[fig:The prompt used for POINTWISE instruction evaluation.]{Figure 5} details the prompt for POINTWISE instruction evaluation. \hyperref[fig:ALPACA fine-tuning data template.]{Figure 6} displays the ALPACA fine-tuning data template.
\label{sec:appendix}

\label{sec:appendix}

\label{sec:appendix}

\label{sec:appendix}

\section{User Interface Display}
\label{sec:appendix-c}
\begin{figure*}[htbp]
    \centering
    \includegraphics[width=\textwidth]{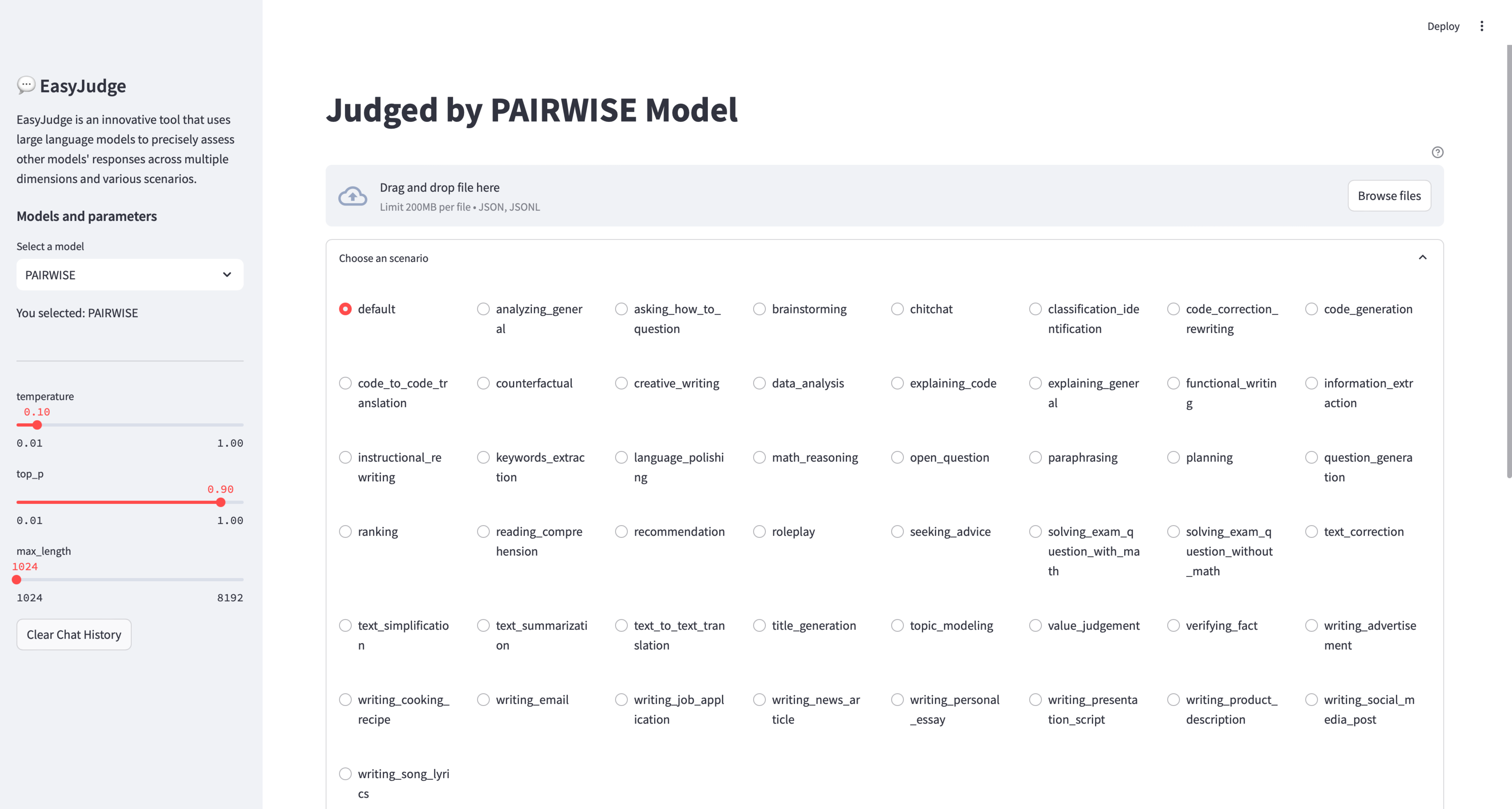}
    \caption{This interface displays all the available scenario for users.}
    \label{fig:user_scenario}
\end{figure*}

\begin{figure*}[htbp]
    \centering
    \includegraphics[width=\textwidth]{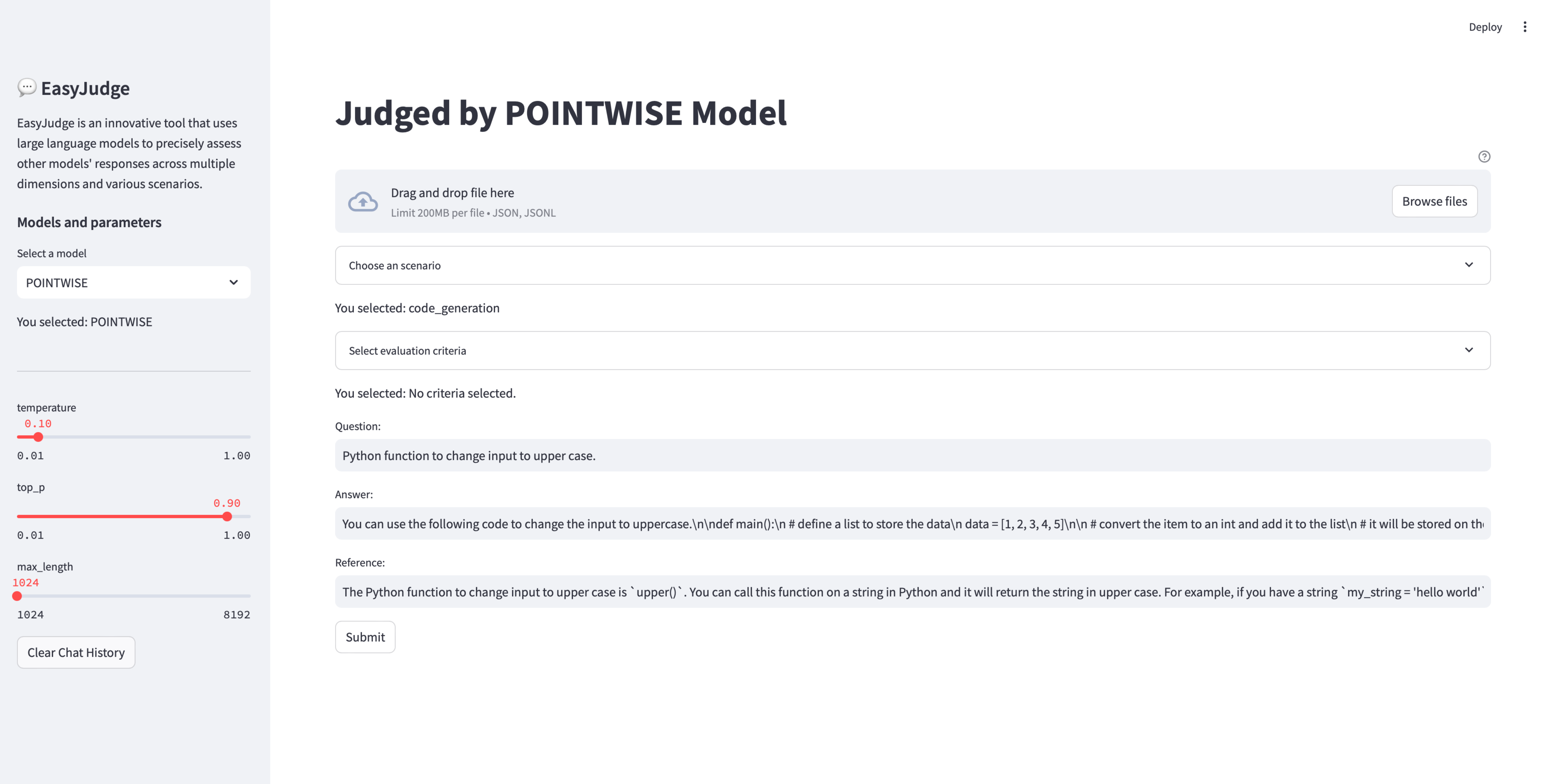}
    \caption{This interface displays one input method for POINTWISE evaluation.}
    \label{fig:user_POINTWISE_1}  
\end{figure*}

\begin{figure*}[htbp]
    \centering
    \includegraphics[width=\textwidth]{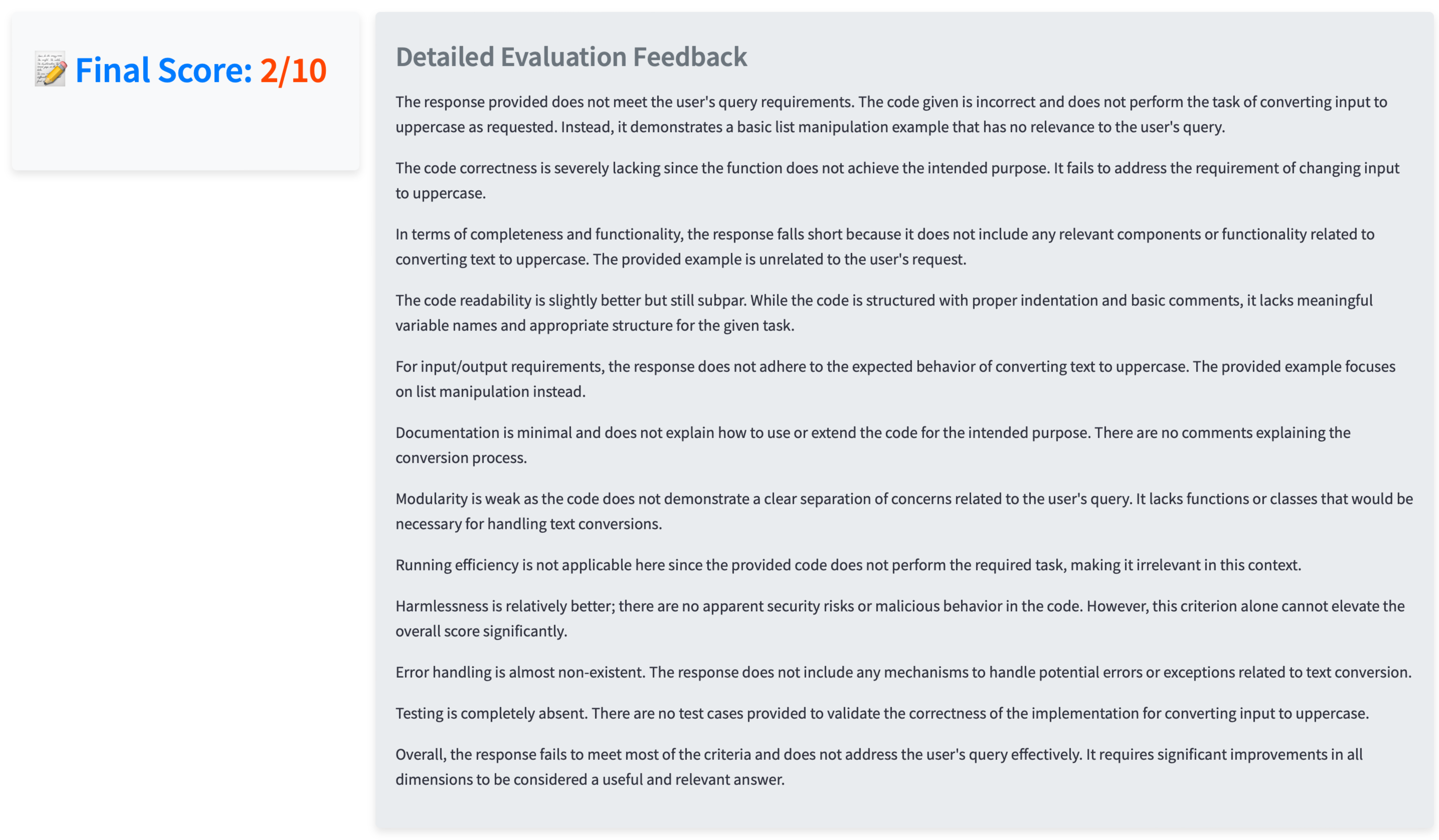}
    \caption{This interface shows an example of detailed feedback for a POINTWISE evaluation.}
    \label{fig:user_POINTWISE_2}  
\end{figure*}

\begin{figure*}[htbp]
    \centering
    \includegraphics[width=\textwidth]{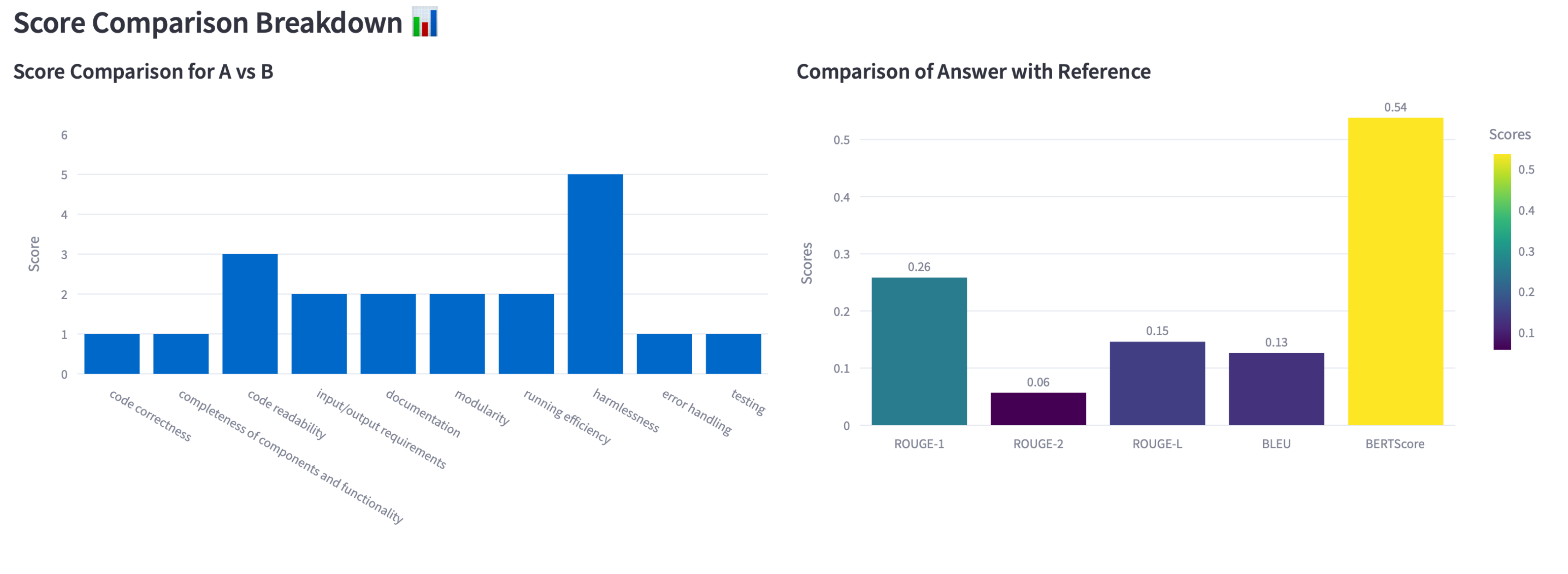}
    \caption{This interface displays a detailed scoring breakdown for a POINTWISE evaluation.}
    \label{fig:user_POINTWISE_3}  
\end{figure*}

This section shows more user interface. \hyperref[fig:user_scenario]{Figure 7} displays all the available scenarios for users, \hyperref[fig:user_POINTWISE_1]{Figure 8} shows one input method for POINTWISE evaluation, \hyperref[fig:user_POINTWISE_2]{Figure 9} presents an example of detailed feedback for a POINTWISE evaluation, and \hyperref[fig:user_POINTWISE_3]{Figure 10} shows a detailed scoring breakdown for a POINTWISE evaluation.
\label{sec:appendix}

\section{Training Details}
\label{sec:appendix-c}
This section shows the parameter settings for training EasyJudge, as shown in \hyperref[tab:training_params]{Table \ref*{tab:training_params}}.

\begin{table*}[htbp]
\centering
\caption{The training parameters of the EasyJudge}
\renewcommand{\thetable}{} 
\begin{tabularx}{\textwidth}{XX}
\toprule
\textbf{Parameters} & \textbf{Value} \\
\midrule
cutoff\_len & 2048 \\
finetuning\_type & lora \\
per\_device\_train\_batch\_size & 4 \\
lr\_scheduler\_type & cosine \\
lora\_rank & 8 \\
lora\_target & q\_proj, v\_proj \\
additional\_target & embed\_tokens, lm\_head, norm \\
learning\_rate & $2\times  10^{-4}$ \\
num train epochs & 1 \\
gradient accumulation steps & 2 \\
max grad norm & 1 \\
lora dropout & 0.05 \\
warmup steps & 0 \\
fp16 & TRUE \\
\bottomrule
\end{tabularx}
\label{tab:training_params}
\end{table*}

\end{document}